\def\year{2016}\relax
\documentclass[letterpaper]{article}
\usepackage{aaai17}
\usepackage{times}
\usepackage{helvet}
\usepackage{courier}
\usepackage{amsfonts,amssymb}
\usepackage{amsmath}
\usepackage{float}
\usepackage{graphicx} 
\usepackage{subfigure}
\newtheorem{theorem}{Theorem}[section]
\newtheorem{lemma}[theorem]{Lemma}

\newtheorem{definition}[theorem]{Definition}

\newtheorem{corollary}[theorem]{Corollary}
\begin{document}
%
\title{Generalization Error Bounds for Optimization Algorithms via Stability}
\author{Qi Meng\textsuperscript{1}, Yue Wang\textsuperscript{2}, Wei Chen\textsuperscript{3}, Taifeng Wang\textsuperscript{3}, Zhi-Ming Ma\textsuperscript{4}, Tie-Yan Liu\textsuperscript{3}\\
	\textsuperscript{1}
	School of Mathematical Sciences, Peking University,
	qimeng13@pku.edu.cn\\\textsuperscript{2}Beijing Jiaotong University, 11271012@bjtu.edu.cn\\\textsuperscript{3}Microsoft Research, $\{$wche, taifengw, tie-yan.liu$\}$@microsoft.com\\\textsuperscript{4}Academy of Mathematics and Systems Science, Chinese Academy of Sciences, mazm@amt.ac.cn
}
\maketitle
\begin{abstract}
	Many machine learning tasks can be formulated as Regularized Empirical Risk Minimization (R-ERM), and solved by optimization algorithms such as gradient descent (GD), stochastic gradient descent (SGD), and stochastic variance reduction (SVRG). Conventional analysis on these optimization algorithms focuses on their convergence rates during the training process, however, people in the machine learning community may care more about the generalization performance of the learned model on unseen test data. In this paper, we investigate on this issue, by using stability as a tool. In particular, we decompose the generalization error for R-ERM, and derive its upper bound for both convex and non-convex cases. In convex cases, we prove that the generalization error can be bounded by the convergence rate of the optimization algorithm and the stability of the R-ERM process, both in expectation (in the order of {\small$\mathcal{O}(1/n)+\mathbb{E}\rho(T))$}, where $\rho(T)$ is the convergence error and $T$ is the number of iterations) and in high probability (in the order of {\small$\mathcal{O}\left(\frac{\log{1/\delta}}{\sqrt{n}}+\rho(T)\right)$} with probability $1-\delta$). For non-convex cases, we can also obtain a similar expected generalization error bound. Our theorems indicate that 1) along with the training process, the generalization error will decrease for all the optimization algorithms under our investigation; 2) Comparatively speaking, SVRG has better generalization ability than GD and SGD. We have conducted experiments on both convex and non-convex problems, and the experimental results verify our theoretical findings.
\end{abstract}

\section{Introduction}

Many machine learning tasks can be formulated as Regularized Empirical Risk Minimization (R-ERM). Specifically, given a training dataset, the goal of R-ERM is to learn a model from a hypothesis space by minimizing the regularized empirical risk defined as the average loss on the training data plus a regularization term.

In most cases, it is hard to achieve an exact minimization of the objective function since the problem might be too complex to have a closed-form solution. Alternatively, we seek an approximate minimization by using some optimization algorithms. Widely used optimization algorithms include the first-order methods such as gradient descent (GD), stochastic gradient descent (SGD), stochastic variance reduction (SVRG) \cite{johnson2013accelerating}, and the second-order methods such as Newton's methods \cite{nocedal2006numerical} and quasi-Newton's methods \cite{nocedal2006numerical}. In this paper, for ease of analysis and without loss of generality, we will take GD, SGD and SVRG as examples. GD calculates the gradient of the objective function at each iteration and updates the model towards the direction of negative gradient by a constant step size. It has been proved that, if the step size is not very large, GD can achieve a linear convergence rate \cite{nesterov2013introductory}. SGD exploits the additive nature of the objective function in R-ERM, and randomly samples an instance at each iteration to calculate the gradient. Due to the variance introduced by stochastic sampling, SGD has to adopt a decreasing step size in order to guarantee the convergence, and the corresponding convergence rate is sublinear in expectation \cite{rakhlin2011making}. In order to reduce the variance in SGD, SVRG divides the optimization process into multiple stages and updates the model towards a direction of the gradient at a randomly sampled instance regularized by a full gradient over all the instances. In this way, SVRG can achieve linear convergence rate in expectation with a constant step size \cite{johnson2013accelerating}.

While the aforementioned convergence analysis can characterize the behaviors of the optimization algorithms in the training process, what the machine learning community cares more is the generalization performance of the learned model on unseen test data. \footnote{Under a related but different setting, i.e., the data instances are successively generated from the underlying distribution, people have proven regret bounds for algorithms like SGD \cite{kakade2009generalization,cesa2004generalization} and SVRG \cite{frostig2015competing}.} As we know, the generalization error of a machine learning algorithm can be decomposed into three parts, the approximation error, the estimation error, and the optimization error. The approximation error is caused by the limited representation power of the hypothesis space $\mathcal{F}$; the estimation error (which measures the difference between the empirical risk and the expected risk) is caused by the limited amount of training data \cite{vapnik1982estimation}\cite{bousquet2002stability}; and the optimization error (which measures the difference between expected risks of the model obtained by the optimization algorithm after $T$ iterations and the true optimum of the regularized empirical risk) is caused by the limited computational power. In \cite{bousquet2008tradeoffs}, Bottou and Bousquet proved generalization error bounds for GD and SGD based on VC-dimension \cite{kearns1999algorithmic}, which unavoidably are very loose in their nature.\footnote{In \cite{hardt2015train}, Hardt $et.al$ studied convex risk minimization via stability, but they did not consider the influence of hypothesis space and the tradeoff between approximation error and estimation error.} The goal of our paper is to develop more general and tighter generalization error bounds for the widely used optimization algorithms in R-ERM.

To this end, we leverage stability \cite{bousquet2002stability} as a tool and obtain the following results:

(1) For convex objective functions, we prove that, the generalization error of an optimization algorithm can be upper bounded by a quantity related to its stability plus its convergence rate in expectation. Specifically, the generalization error bound is in the order of {\small$\mathcal{O}(1/n+\mathbb{E}\rho(T))$}, where $\rho(T)$ is the optimization convergence error and $T$ is the number of iterations. This indicates that along with the optimization process on the training data, the generalization error will decrease, which is consistent with our intuition.

(2) For convex objective functions, we can also obtain a high probability bound for the generalization error. In particular, the bound is in the order of {\small$\mathcal{O}\left(\frac{\log{1/\delta}}{\sqrt{n}}+\rho(T)\right)$} with probability at least {\small$1-\delta$}. That is, if an algorithm has a high-probability convergence bound, we can get a high-probability generalization error bound too, and our bound is sharper than those derived in the previous literature.

(3) Based on our theorems, we analyze the time for different optimization algorithms to achieve the same generalization error, given the same amount of training data. We find that SVRG outperforms GD and SGD in most cases, and although SGD can quickly reduce the test error at the beginning of the training process, it slows down due to the decreasing step size and can hardly obtain the same test error as GD and SVRG when $n$ is large.

(4) Some of our theoretical results can be extended to the nonconvex objective functions, with some additional assumptions on the distance between the global minimizer and the stationary local minimizers.

We have conducted experiments on linear regression, logistic regression, and fully-connected neural networks to verify our theoretical findings. The experimental results are consistent with our theory: (1) when the training process goes on, the test error decreases; (2) in most cases, SVRG has better generalization performance than GD and SGD.


\section{Preliminaries}\label{pre}
In this section, we briefly introduce the R-ERM problem, and popular optimization algorithms to solve it.

\subsection{R-ERM and its Stability}
Suppose that we have a training set
$S=\{z_1=(x_1,y_1),...,z_n=(x_n,y_n)\}$ with $n$ instances that are i.i.d. sampled from $\mathcal{Z=\mathcal{X}\times\mathcal{Y}}$ according to an unknown distribution $\mathcal{P}$. The goal is to learn a good prediction model $f\in\mathcal{F}: \mathcal{X}\to\mathcal{Y}$, whose prediction accuracy at instance $(x,y)$ is measured by a loss function $l(y,f(x))=l(f,z)$. Different learning tasks may use different loss functions, such as the least square loss {\small$(f(x)-y)^2$} for regression, and the logistic loss {\small$\log{(1+e^{-yf(x)})}$} for classification. We learn the prediction model from the training set $S$, and will use this model to give predictions for unseen test data.

R-ERM is a very common way to achieve the above goal. Given loss function $l(f,z)$, we aim to learn a model $f^*$ that minimizes the expected risk {\small$$R(f)=\mathbb{E}_{z\sim\mathcal{P}}l(f,z).$$} Because the underlying distribution $\mathcal{P}$ is unknown, in practice, we learn the prediction model by minimizing the regularized empirical risk over the training instances, which is defined as below,
{\small\begin{equation}\label{nota2}
	R_{S}^r(f)=\frac{1}{n}\sum_{i=1}^nl(f,z_i)+\lambda N(f).
	\end{equation}}
Here, the regularization term $\lambda N(f)$ helps to restrict the capacity of the hypothesis space $\mathcal{F}$ to avoid overfitting. In this paper, we consider $N(f)$ as a norm in a reproducing kernel Hilbert space (RKHS): $N(f)=\|f\|_k^2$ where $k$ refers to the kernel \cite{wahba2000introduction}.

As aforementioned, our goal is expected risk minimization but what we can do in practice is empirical risk minimization instead. The gap between these two goals is measured by the so-called estimation error, which is usually expressed in the following way: the expected risk is upper bounded by the empirical risk plus a quantity related to the capacity of the hypothesis space  \cite{vapnik1982estimation}\cite{bousquet2008tradeoffs}. One can choose different ways to measure the capacity of the hypothesis space, and stability is one of them, which is proved to be able to produce tighter estimation error bound than VC dimension \cite{kearns1999algorithmic}. There has been a venerable line of research on estimation error analysis based on stability, dated back more than thirty years ago \cite{bousquet2002stability,devroye1979distribution,kearns1999algorithmic,mukherjee2006learning,shalev2010learnability}. The landmark work by Bousquet and Elisseeff \cite{bousquet2002stability} introduced the following definitions of uniform loss stability and output stability.
\begin{definition}(Uniform Loss Stability)\label{def1}
	An algorithm $A$ has uniform stability $\beta_0$ with respect to loss function $l$ if the following holds $\forall S\in\mathcal{Z}^n,\forall j\in\{1,\cdots,n\},$
	{\small\begin{equation}\label{stability}
		\left|\mathbb{E}_A\left[l(A_{S},\cdot)\right]-\mathbb{E}_A\left[l(A_{S^{\setminus j}},\cdot)\right]\right|\leq \beta_0,
		\end{equation}}
	where $A_S, A_{S^{\setminus j}}$ are the outputs of algorithm $A$ based on $S$ and $S^{\setminus j}=\{z_1,\cdots,z_{j-1},z_{j+1},\cdots,z_n\}$, respectively.
\end{definition}
\begin{definition}\label{def2}(Output Stability)
	An algorithm has output stability $\beta_1$ if the following holds $\forall S\in\mathcal{Z}^n,\forall j\in\{1,\cdots,n\},$
	{\small\begin{equation}
		\|A_{S}-A_{S^{\setminus j}}\|_{\mathcal{F}_c}\leq\beta_1,
		\end{equation}}
	where $\|\cdot\|_{\mathcal{F}_c}$ denotes the norm in hypothesis space $\mathcal{F}_c$.
\end{definition}

From the above definitions, we can see that stability measures the change of the loss function or the produced model of a given learning algorithm if one instance in the training set is changed. For example, if the loss function is convex and $L$-Lipschitz w.r.t. $f$, the corresponding R-ERM algorithm with regularization term $N(f)=\|f\|_k^2$ has stability $\beta_0\leq \frac{L^2 K^2}{2\lambda n}$ and $\beta_1\leq\frac{LK}{2\lambda n}$, where $K$ is the upper bound of the kernel norm \cite{bousquet2002stability}.

\begin{small}
	\begin{table*}[t]
		\centering
		\begin{tabular}{l|c|c|c|cc} \hline
			&Convex&Convex&Nonconvex&Nonconvex&\\
			&Number of iterations&Number of data passes&Number of iterations&Number of data passes&\\
			\hline
			GD&$\mathcal{O}(\kappa\log{(1/\epsilon)})$&$\mathcal{O}\left(n\kappa\ln{(1/\epsilon)}\right)$&$\mathcal{O}(1/\epsilon)$&$\mathcal{O}\left(n(1/\epsilon)\right)$\\
			SGD&$\mathcal{O}(\kappa^2/\epsilon)$&$\mathcal{O}\left(\kappa^2/\epsilon\right)$&$\mathcal{O}(1/\epsilon^2)$&$\mathcal{O}\left(1/\epsilon^2\right)$\\
			SVRG&$\mathcal{O}(\kappa\log{(1/\epsilon)})$&$\mathcal{O}\left(n+\kappa\ln{(1/\epsilon)}\right)$&$\mathcal{O}(1/\epsilon)$&$\mathcal{O}\left(n+n^{2/3}(1/\epsilon)\right)$\\
			\hline
		\end{tabular}
		\caption{Convergence rate of GD, SGD, SVRG in convex and nonconvex cases, where $\epsilon$ is the targeted accuracy, $\kappa$ is the condition number.}
		\label{convergence convex}
	\end{table*}
\end{small}
\subsection{Optimization Algorithms}
Many optimization methods can be used to solve the R-ERM problem, including the first-order methods such as Gradient Descent (GD) \cite{nesterov2013introductory}, Stochastic Gradient Descent (SGD) \cite{rakhlin2011making}, and Stochastic Variance Reduction (SVRG) \cite{johnson2013accelerating}, as well as the second-order methods such as Newton's methods \cite{nocedal2006numerical} and quasi-Newton's methods \cite{byrd2016stochastic}. We will take the first-order methods as examples in this paper, although many of our analysis can be easily extended to other optimization algorithms.

Let us consider model $f$ parameterized by $w$. The update rules of GD, SGD, and SVRG are summarized as follows.
\textbf{Gradient Descent (GD)}
{\small\begin{equation}
w_{t+1}=w_t-\eta\nabla R_{S}^r(w_t).
\end{equation}}
\textbf{Stochastic Gradient Descent (SGD)}
{\small\begin{equation}
w_{t+1}=w_t-\eta_tg(w_t) .
\end{equation}}
\textbf{Stochastic Variance Reduced Gradient (SVRG)}
{\small\begin{eqnarray}
		&&v_s^t=g(w_s^t)-\nabla R_{S}^r(w_s^t)+\nabla R_{S}^r(\tilde{w}^{t-1})\\
		&&w_{s+1}^t=w_s^t-\eta v_s^t.
	\end{eqnarray}}
 where $g(\cdot)$ is the gradient of $\nabla R_{S}^r(\cdot)$ at randomly sampled training instances, $w_s^t$ is the output parameter at the $s$-th iteration in the $t$-th stage, and $\tilde{w}^{t-1}$ is the final output in stage {\small$t-1$}.

When the loss function is strongly convex and smooth with respect to the model parameters, GD can achieve linear convergence rate; SGD can only achieve sublinear convergence rate due to the variance introduced by stochastic sampling (but in each iteration, it only needs to compute the gradient over one instance and thus can be much faster in speed); SVRG can achieve linear convergence rate by reducing the variance and in most iterations it only needs to compute the gradient over one instance. \footnote{The second-order methods can get quadratic convergence rate \cite{nocedal2006numerical}. However, as compared with the first-order methods, the computation complexity of the second-order methods could be much higher due to the calculation of the second-order information.} When the loss functions are nonconvex w.r.t. the model parameters (e.g., neural networks), GD \cite{nesterov2013introductory}, SGD \cite{ghadimi2013stochastic}, and SVRG \cite{reddi2016stochastic} still have convergence properties (although regarding a different measure of convergence). For ease of reference, we summarize the convergence rates of the aforementioned optimization algorithms in both convex and nonconvex cases in Table \ref{convergence convex}.

	\section{Generalization Analysis}\label{sec3}
	In this section, we will analyze the generalization error for optimization algorithms by using stability as a tool. Firstly, we introduce the definition of generalization error and its decomposition. Then, we prove the generalization error bounds of optimization algorithms in both convex and nonconvex cases. The proof details of all the lemmas and theorems are placed in the supplementary materials due to space limitation.
	\subsection{Generalization Error and its Decomposition}
	As we mentioned in Section \ref{pre}, R-ERM minimizes the regularized empirical risk, i.e.,
	{\small\begin{equation}
		f_{S,r}^*:=argmin_{f\in\mathcal{F}}R_{S}^r(f)
		\end{equation}}
 as an approximation of the expected risk minimization:
 	{\small\begin{equation}
 		f^{*}:=argmin_f R(f).
 		\end{equation}}
Denote the empirical risk {\small$R_S(f)=\frac{1}{n}\sum_{i=1}^nl(f,z_i)$}. It is clear that, the minimization of {\small$R_S^r(f)$} in $\mathcal{F}$ is equivalent to the minimization of $R_S(f)$ in {\small$\mathcal{F}_c=\{f\in\mathcal{F},N(f)\leq c\}$} for some constant $c$. That is,
\begin{equation}\label{regularization_eqn}
f_{S,r}^*=f_{S,\mathcal{F}_c}^*:=argmin_{f\in\mathcal{F}_c} R_S(f).
\end{equation}

Denote the minimizer of the expected risk $R(f)$ in the hypothesis space $\mathcal{F}_c$ as  {\small$f_{\mathcal{F}_c}^* $},  i.e.,
 {\small\begin{equation}
 	f_{\mathcal{F}_c}^*:=argmin_{f\in\mathcal{F}_c}R(f).
 	\end{equation}}
In many practical cases, neither $f_{S,r}^*$ nor $f_{S,\mathcal{F}_c}^*$ has a closed form. What people do is to implement an iterative optimization algorithm $A$ to produce the prediction model. We denote the output model of algorithm $A$ at iteration $T$ over $n$ training instances as $f_T(A,n,\mathcal{F}_c)$. We use \emph{generalization error} to denote the difference between the expected risk of this learnt model and the optimal expected risk, as follows,
\begin{equation}
\mathcal{E}(A,n,\mathcal{F}_c,T)=R(f_T(A,n,\mathcal{F}_c))-R(f^*).
\end{equation}

As known, the generalization error can be decomposed into the three components,
	{\small\begin{eqnarray}
		&&\mathcal{E}(A,n,\mathcal{F}_c,T)\\
		&=&R(f_T)-R(f_{S,\mathcal{F}_c}^{*})+R(f_{S,\mathcal{F}_c}^{*})-R(f_{\mathcal{F}_c}^*)\\
		&\quad&+R(f_{\mathcal{F}_c}^{*})-R(f^*) \nonumber\\
		&:=&\mathcal{E}_{opt}(A,n,\mathcal{F}_c,T)+\mathcal{E}_{est}(n,\mathcal{F}_c)+\mathcal{E}_{app}(\mathcal{F}_c).
		\end{eqnarray}}	
The item {\small$\mathcal{E}_{app}(\mathcal{F}_c):=R(f_{\mathcal{F}_c}^*)-R(f^*)$}, is called \emph{approximation error}, which is caused by the limited representation power of the hypothesis space $\mathcal{F}_{c}$. With the hypothesis space increasing, (i.e., $c$ is increasing), the approximation error will decrease. The item {\small$\mathcal{E}_{est}(n,\mathcal{F}_c):= R(f_{S,\mathcal{F}_c}^{*})-R(f_{\mathcal{F}_c}^*)$}, is called \emph{estimation error}, which is caused by the limited amount of the training data (which leads to the gap between the empirical risk and the expected risk). It will decrease with the increasing training data size $n$, and the decreasing capacity of the hypothesis space $\mathcal{F}_c$. The item {\small$\mathcal{E}_{opt}(A,n,\mathcal{F}_c,T):=R(f_T)-R(f_{S,\mathcal{F}_c}^*)$}, is called \emph{optimization error}, which measures the sub-optimality of the optimization algorithms in terms of the expected risk. It is caused by the limited computational resources. \footnote{For simplicity, we sometimes denote $f_T(A,n,\mathcal{F}_c)$, $\mathcal{E}(A,n,\mathcal{F}_c,T)$, $\mathcal{E}_{app}(\mathcal{F}_c)$, $\mathcal{E}_{est}(n,\mathcal{F}_c)$, $\mathcal{E}_{opt}(A,n,\mathcal{F}_c,T)$ as $f_T$, $\mathcal{E}$, $\mathcal{E}_{app}$, $\mathcal{E}_{est}$, $\mathcal{E}_{opt}$, respectively.}

Please note that, the optimization error under our study differs from the target in the conventional convergence analysis of optimization algorithms. In the optimization community, the following two objectives
{\small\begin{eqnarray}
\rho_0(T)=R_S(f_T)-R_S(f_{S,\mathcal{F}_c}^{*}) ; \quad\rho_1(T)=\|f_T-f_{S,\mathcal{F}_c}^{*}\|_{\mathcal{F}_c}^2 \label{eqcon}
\end{eqnarray}}
are commonly used in convex cases, and
{\small\begin{equation}\label{eqnon}
\rho_2(T)=\|\nabla R_S^r(f_T)\|^2
\end{equation}}
is commonly used in nonconvex cases.
To avoid confusion, we call them \emph{convergence error} and their corresponding upper bounds \emph{convergence error bounds}. Please note although convergence error is different from optimization error, having a convergence error bound plays an important role in guaranteeing a generalization error bound. In the following subsections, we will prove the generalization error bound for typical optimization algorithms, by using the stability techniques, based on their convergence error bounds.

		\subsection{Expected Generalization Bounds for Convex Case}
	The following theorem gives an expected generalization error bounds in the convex case.
	
	 	\begin{theorem}\label{thm1}
		Consider an R-ERM problem, if the loss function is $L$-Lipschitz continuous, $\gamma$-smooth, and convex with respect to the prediction output vector, we have {\small\begin{eqnarray}
				\mathbb{E}_{S,A}\mathcal{E}&\leq&\mathcal{E}_{app}+2\beta_0+\mathbb{E}_{S,A}\rho_0(T)+\frac{\gamma\mathbb{E}_{S,A}\rho_1(T)}{2}\nonumber\\
				&&+\sqrt{\mathbb{E}_{S,A}\rho_1(T)\left(\frac{L^2}{2n}+6L\gamma\beta_1\right)},
				\end{eqnarray}}
where $\beta_0,\beta_1$ are the uniform stability and output stability of the R-ERM process as defined in \ref{def1} and \ref{def2}, $\rho_0(T)$ and $\rho_1(T)$ are the convergence errors defined in Eqn \ref{eqcon}.
		\end{theorem}
	
From Theorem \ref{thm1}, we can see that the generalization error can be upper bounded by the stability $\beta_0$ and $\beta_1$, the convergence errors of the optimization algorithms $\rho_0(T)$ and $\rho_1(T)$, and the well-studied approximation error \cite{vapnik1998statistical}. As the training process goes on, both $\mathbb{E}\rho_0(T)$ and $\mathbb{E}\rho_1(T)$ will decrease. Therefore, the expected generalization error will decrease too. This is consistent with our intuition. Better optimizations will lead to better expected generalization performance.

In order to prove Theorem \ref{thm1}, we need the following two lemmas, whose proofs are placed in the supplementary materials due to space restrictions.

		\begin{lemma}\label{lemma}
			For R-ERM problems, we have $\forall j\in\{1,\cdots,n\}$:
			{\small\begin{equation}
				\mathbb{E}_S\left[R(f_{S,\mathcal{F}_c}^*)-R_{S}(f_{S,\mathcal{F}_c}^*)\right]=\mathbb{E}_S\left[l(f_{S,\mathcal{F}_c}^*,z'_j)-l(f_{S^{j},\mathcal{F}_c}^*,z'_j)\right]\label{eq7}
				\end{equation}} and
				{\small\begin{equation}
			\mathbb{E}_S[\nabla R(f_{S,\mathcal{F}_c}^*)-\nabla R_{S}(f_{S,\mathcal{F}_c}^*)]=\mathbb{E}_S[\nabla_fl(f_{S,\mathcal{F}_c}^*,z'_j)-\nabla_fl(f_{S^{j},\mathcal{F}_c}^*,z'_j)]\label{eq8},
				\end{equation}}
			where $S^j=\{z_1,\cdots,z_{j-1},z'_j,z_{j+1},\cdots,z_n\}$, and $f_{S^j,\mathcal{F}_c}^*$ is the minimizer of $R_{S^j}(f)$ in $\mathcal{F}_c$ .
		\end{lemma}
			
			\begin{lemma}\label{lemma1}
				Assume that the loss function is $L$-Lipschitz and $\gamma$-smooth w.r.t. the prediction output vector, we have
				{\small\begin{equation}
					\mathbb{E}_S[\nabla R(f_{S,\mathcal{F}_c}^*)-\nabla R_S(f_{S,\mathcal{F}_c}^*)]^2\leq\frac{L^2}{2n}+6L\gamma\beta_1.
					\end{equation}}
			\end{lemma}
	\textit{Proof Sketch of Theorem \ref{thm1}:}
	
	\textit{Step 1:}
	Since the loss function is convex and $\gamma$-smooth w.r.t. $f$, we can get that $R(f)$ is $\gamma$-smooth and $R_S(f)$ is convex w.r.t $f$. We decompose $\mathcal{E}_{opt}$ as below:
	{\small\begin{eqnarray*}
	\mathcal{E}_{opt}&\leq&\left(\nabla R(f_{S,\mathcal{F}_c}^*)-\nabla R_{S}(f_{S,\mathcal{F}_c}^*)\right)^T(f_T-f_{S,\mathcal{F}_c}^*) \\
	&\quad&+R_{S}(f_T)-R_{S}(f_{S,\mathcal{F}_c}^*)+\frac{\gamma}{2}\|f_T-f_{S,\mathcal{F}_c}^*\|_{\mathcal{F}_c}^2, \nonumber
	\end{eqnarray*}}
	We can use {\small$\rho_0(T)$}, {\small$\rho_1(T)$} and Lemma \ref{lemma1}, to get an upper bound of $\mathbb{E}_{S,A}\mathcal{E}_{opt}$.
	
	\textit{Step 2:}
    Since $R_S(f_{S,\mathcal{F}_c}^*)\leq R_S(f_{\mathcal{F}_c}^*)$, we have   {\small\begin{equation*}\mathcal{E}_{est}\leq\left[R(f_{S,\mathcal{F}_c}^*)-R_S(f_{S,\mathcal{F}_c}^*)\right]+\left[R_S(f_{\mathcal{F}_c}^*)-R(f_{\mathcal{F}_c}^*)\right].\end{equation*}}
    We have {\small$\mathbb{E}_S\left[R_S(f_{\mathcal{F}_c}^*)-R(f_{\mathcal{F}_c}^*)\right]=0$}. By using Lemma \ref{lemma}, we can bound $\mathbb{E}_{S}\mathcal{E}_{est}$.
    By combining the upper bounds of $\mathbb{E}_{S,A}\mathcal{E}_{opt}$ and $\mathbb{E}_{S}\mathcal{E}_{opt}$, we can get the results.

After proving the general theorem, we consider a special case - an R-ERM problem with kernel regularization term $\lambda\|f\|_k^2$. In this case, we can derive the concrete expressions of the stability and convergence error. In particular, {\small$\beta_0=\mathcal{O}(1/\lambda n)$, $\beta_1=\mathcal{O}(1/\lambda n)$} and $\rho_1(T)$ is equivalent to $\|w_T-w_{S,r}^*\|^2$. If the loss function is convex and smooth w.r.t. parameter $w$, $R_S^r(w)$ with $N(f)=\|f\|_k^2$ is strongly convex and smooth w.r.t $w$. In this case, $\rho_0(T)$ dominates $\rho_1(T)$, i.e., $\rho_0(T)$ is larger than $\rho_1(T)$ w.r.t the order of $T$. Therefore, we can obtain the following corollary.
		
		\begin{corollary}
			For an R-ERM problem with a regularization term $\lambda\|f\|_k^2$, under the same assumptions in Theorem \ref{thm1}, and further assuming that the loss function is convex and smooth w.r.t parameter $w$, we have
			{\small\begin{eqnarray}
					\mathbb{E}_{S,A}\mathcal{E}\leq \mathcal{E}_{app}+\mathcal{O}\left(\frac{1}{\lambda n}+\mathbb{E}_{S,A}\rho_0 (T)\right).
				\end{eqnarray}}
			\end{corollary}

		\subsection{High-Probability Generalization Bounds for Convex Case}
The following theorem gives a high-probability bound of $\mathcal{E}$ in the convex case. Due to space limitation, we put the proof in the supplementary materials.
		\begin{theorem}\label{high}
For an R-ERM problem, if the loss function is L-Lipschitz continuous, $\gamma$-smooth and convex with respect to the prediction output vector, and $0\leq l(f_{S,\mathcal{F}_c}^*,z)\leq M$ for arbitrary $z\in\mathcal{Z}$ and $S\in\mathcal{Z}^n$, then with probability at least $1-\delta$, we have
			{\small\begin{eqnarray*}
				\mathcal{E}&\leq&\mathcal{E}_{app}+2\beta_0+\rho_0(T)+\frac{\gamma}{2}\rho_1(T)+2\gamma\beta_1\sqrt{\rho_1(T)}\\	
				&&+\left(4n\beta_0+2M+(4n\gamma\beta_1+L)\sqrt{\rho_1(T)}\right)\sqrt{\frac{\ln{4/\delta}}{2n}}.
				\end{eqnarray*}}
		\end{theorem}
The high-probability bound is consistent with the expected bound given in the previous subsection. That is, the high-probability generalization bound will also decrease along with the training process. In addition, we can also get a corollary for the special case of R-ERM with kernel regularization.

			\begin{corollary}\label{coro}
				For an R-ERM problem with kernel regularization term $\lambda\|f\|_k^2$, under the same assumptions in Theorem \ref{high}, and further assuming that the loss function is convex and smooth w.r.t parameter $w$,  we have, with probability at least $1-\delta$,
				{\small\begin{eqnarray*}
						\mathcal{E}&\leq& \mathcal{E}_{app}+\mathcal{O}\left(\sqrt{\frac{\log 1/\delta}{n}}+\rho_0(T)\right).
					\end{eqnarray*}}
				\end{corollary}

Rakhlin \textit{et.al.} \cite{rakhlin2011making} proved a high-probability convergence rate for SGD. For GD, the training process is deterministic. By plugging the order of $\beta_0$ and $\beta_1$ in SGD and GD, we have the following corollary.

			\begin{corollary}
				For an R-ERM problem with kernel regularization, under the assumptions in Corollary \ref{coro}, with probability at least $1-\delta$, the generalization error of SGD and GD can be upper bounded as follows,
				{\small\begin{eqnarray*}
					\mathcal{E}_{SGD}&\leq&\mathcal{E}_{app}+\mathcal{O}\left(\sqrt{\frac{\ln{1/\delta}}{n}}\right)+\mathcal{O}\left(\frac{\kappa^2\log(\log (T)/\delta)}{T}\right);\\
\mathcal{E}_{GD}&\leq&\mathcal{E}_{app}+\mathcal{O}\left(\sqrt{\frac{\ln{1/\delta}}{n}}\right)+\mathcal{O}\left(e^{-\kappa T}\right),
					\end{eqnarray*}}
where $\kappa$ is the condition number.
				
			\end{corollary}
	
				\subsection{Expected Generalization Bounds for Nonconvex Case}
				In this subsection, we consider the case in which the loss function is convex w.r.t. the prediction output vector, but non-convex w.r.t. the model parameter. This case can cover deep neural networks, which are state-of-the-art AI techniques nowadays.

For the non-convex case, the definition of convergence error is a little different, as shown by Eq. (\ref{eqnon}).
It measures whether the solution is close to a critical point, which is defined and further categorized as follows.

				\begin{definition}
					Consider the objective $R_S^r$ and parameter $w$. If $\nabla R_{S}^r(w)=0$, we say $w$ is a critical point of $R_S^r$; if $\nabla R_{S}^r(w)$ has at least one strictly negative eigenvalue, we say $w$ is a strict saddle point. If each critical point $w$ is either a local minimum or a strict saddle point, we say that $R_{S}^r$ satisfies the strict saddle property.
				\end{definition}

The following theorem gives the expected generalization error bound for non-convex cases under the widely used assumptions \cite{lian2015asynchronous,reddi2016stochastic,lee2016gradient}.
					
				\begin{small}
					\begin{table*}[t!]
						\centering
						\begin{tabular}{l|c|c|c|c|cc} \hline
							&Cost&Convex&Convex&Nonconvex&Nonconvex&\\
							&per Iteration&Iterations&Time&Iterations&Time&\\
							\hline
							GD&{\small$\mathcal{O}(nd)$}&{\small$\mathcal{O}\left(\kappa\ln{n}\right)$}&{\small$\mathcal{O}\left(nd\kappa\ln{n}\right)$}&{\small$\mathcal{O}\left(1/\epsilon_0^2+n^2\right)$}&{\small$\mathcal{O}\left(n/\epsilon_0^2+n^3\right)$}\\
							SGD&{\small$\mathcal{O}(d)$}&{\small$\mathcal{O}\left(\kappa^2n\right)$}&{\small$\mathcal{O}\left(nd\kappa^2\right)$}&{\small$\mathcal{O}\left(1/\epsilon_0^4+n^4\right)$}&{\small$\mathcal{O}\left(1/\epsilon_0^4+n^4\right)$}\\
							SVRG&{\small$\mathcal{O}(d)$}&{\small$\mathcal{O}\left(\kappa\ln{n\kappa}\right)$}&{\small$\mathcal{O}\left((nd+d\kappa)\ln{n\kappa}\right)$}&{\small$\mathcal{O}\left(1/\epsilon_0^2+n^2\right)$}&{\small$\mathcal{O}(n^{2/3}/\epsilon_0^2+n^{8/3})$}\\
							\hline
						\end{tabular}
						\caption{Sufficient training iteration/time for convex and nonconvex case}
						\label{sufficient training time}
					\end{table*}
				\end{small}

				\begin{theorem}\label{non}
					If $R_{S}^r$ is $\mu$-strongly convex in the $\epsilon_0$- neighborhood of arbitrary local minimum $w_{loc}$, satisfies strict saddle point property, $L$- Lipschitz continuous, $\gamma$-smooth and continuously twice differential w.r.t the model parameter $w$, and the loss function is convex w.r.t $f$, then we have
					{\small\begin{equation*}
						\mathbb{E}_{S,A}\mathcal{E}\leq\mathcal{E}_{app}+2\beta_0+R(w_{loc})-R(w_{S,\mathcal{F}_c}^*)+\frac{L}{\mu}\sqrt{\min_{t=1,\cdots,T}\mathbb{E}_{S,A}\rho_2(t)},
						\end{equation*}}
					where {\small$T\geq T_1$} and {\small$T_1$} is the number of iterations to achieve  {\small$\min_{t=1,\cdots,T_1}\mathbb{E}_{S,A}\left[\rho_2(t)\right]\leq\gamma^2\epsilon_0^2$}.
				\end{theorem}

Similarly to the convex case, from the above theorem we can see that with the training process going on, the generalization error in the nonconvex case will also decrease. In addition, we can also derive specific bound for the R-ERM with kernel regularization.

				\section{Sufficient Training and Optimal Generalization Error}\label{sec4}
In this section, we make further discussions on the generalization bound. In particular, we will explore the sufficient training iterations, and the optimal generalization error given the training data size.

As shown in Section \ref{sec3}, the generalization error bounds consist of an estimation error related to the training data size $n$ and an optimization error related to the training iteration $T$. Given a machine learning task with fixed training size $n$, at the early stage of the training process (i.e., $T$ is relatively small), the optimization error will dominate the generalization error; when $T$ becomes larger than a threshold, the optimization error will decrease to be smaller than the estimation error (i.e. $\mathcal{O}(1/n)$), and then the estimation error will dominate the generalization error. We call this threshold \emph{sufficient training iteration} and the corresponding training time \emph{sufficient training time}. The generalization error with the optimization algorithm sufficiently trained is called \emph{optimal generalization error}. Given the generalization error bound, we can derive the sufficient training iteration/time. For ease of analysis, we list the sufficient training iteration/time of GD, SGD, and SVRG for both convex and nonconvex cases in Table \ref{sufficient training time}.
				
From Table \ref{sufficient training time}, we have the following observations. For the convex case, when the condition number $\kappa$  is much smaller than $n$, GD, SGD and SVRG have no big differences from each other in their sufficient training iterations; when $\kappa$ is comparable with $n$, e.g., $\kappa=\mathcal{O}(\sqrt{n})$, \footnote{In some cases, $\kappa$ is related to the regularization coefficient $\lambda$ and $\lambda$ is determined by the data size $n$ \cite{vapnik1998statistical}\cite{shamir2014communication}.} the sufficient training time for GD, SGD and SVRG is {\small$\mathcal{O}(n\sqrt{n}d\ln{n})$, $\mathcal{O}(n^2d)$, {\small$\mathcal{O}(nd\ln{n})$}}, respectively. That is, SVRG corresponds to a shorter sufficient training time than GD and SVRG. For the non-convex case, if {\small$\epsilon_0 \leq\mathcal{O}(1/n)$}, which is more likely to happen for small data size $n$, the first term in the sufficient training time dominates, and it is fine to terminate the training process at {\small$T=T_1$}. SVRG requires shorter training time than GD and SGD by at least an order of {\small$\mathcal{O}(n^{1/3})$} and {\small$\mathcal{O}(n^{4/3})$}, respectively. If $\epsilon_0$ is larger than $\mathcal{O}(1/n)$, which is more likely to happen for large data size $n$, the sufficient training time for GD, SGD, and SVRG is {\small$\mathcal{O}(n^3)$, $\mathcal{O}(n^4)$, and $ \mathcal{O}(n^{8/3})$}, respectively. In this case, SVRG requires shorter training time than GD and SGD by an order of {\small$\mathcal{O}(n^{1/3})$} and {\small$\mathcal{O}(n^{4/3})$}, respectively.

\section{Experiments}
\begin{figure}[ht]
	\centering
		\subfigure[]{
		\label{figb}
		\includegraphics[width=1.02in]{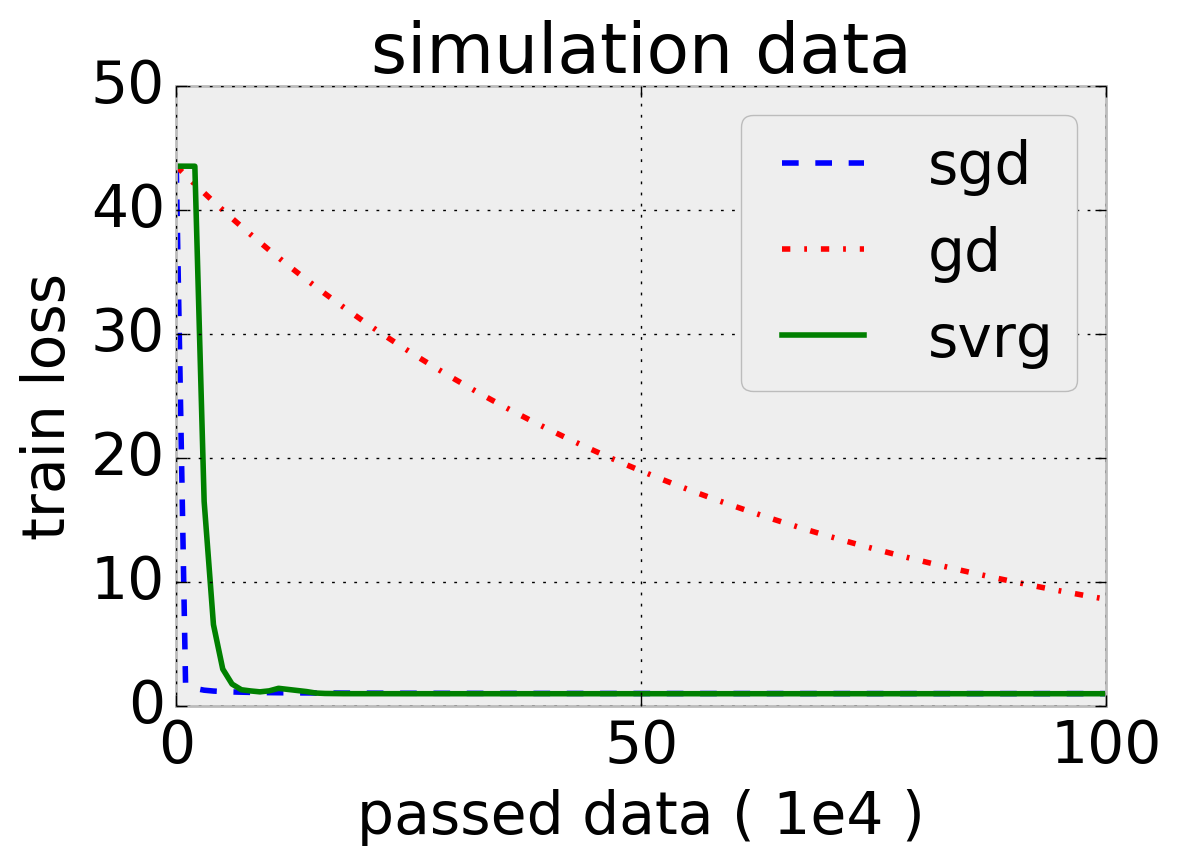}}
	\subfigure[]{
		\label{figc}
		\includegraphics[width=1.02in]{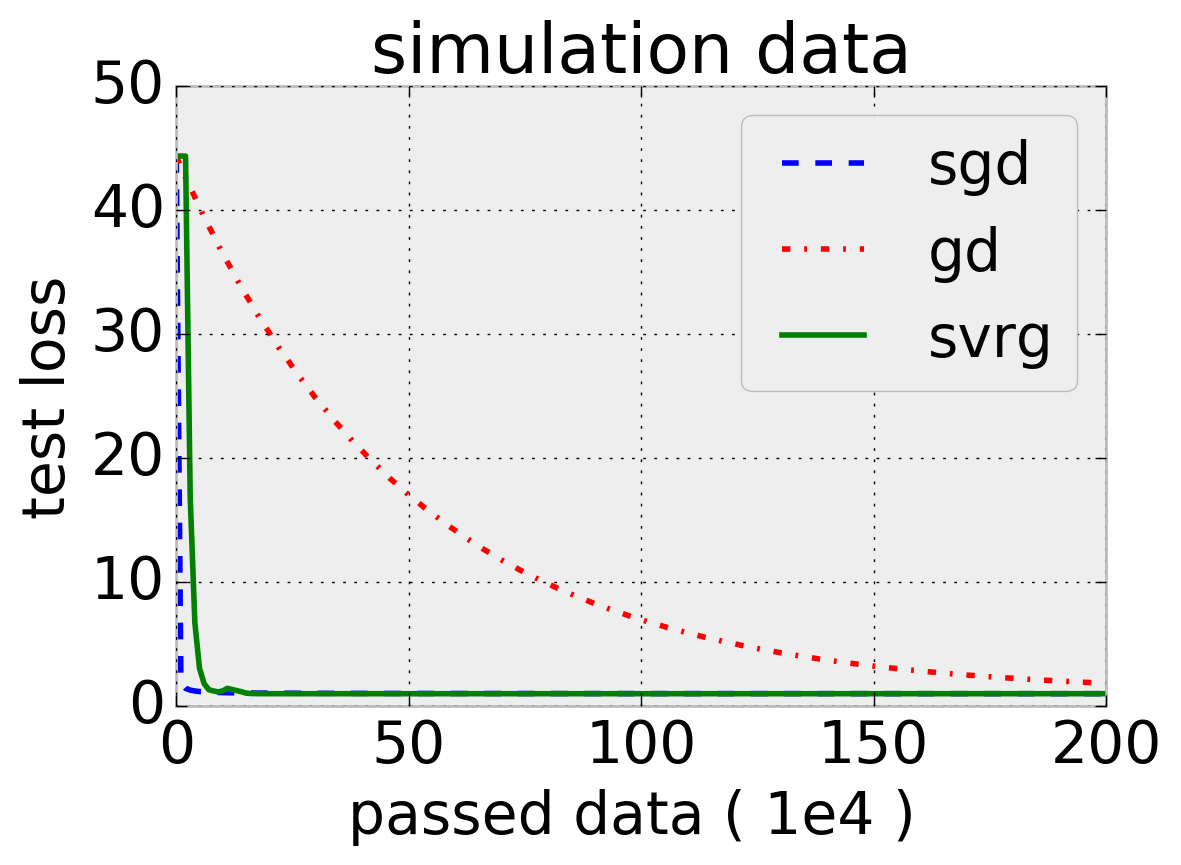}}
	\subfigure[]{
		\label{figa}
		\includegraphics[width=1.02in]{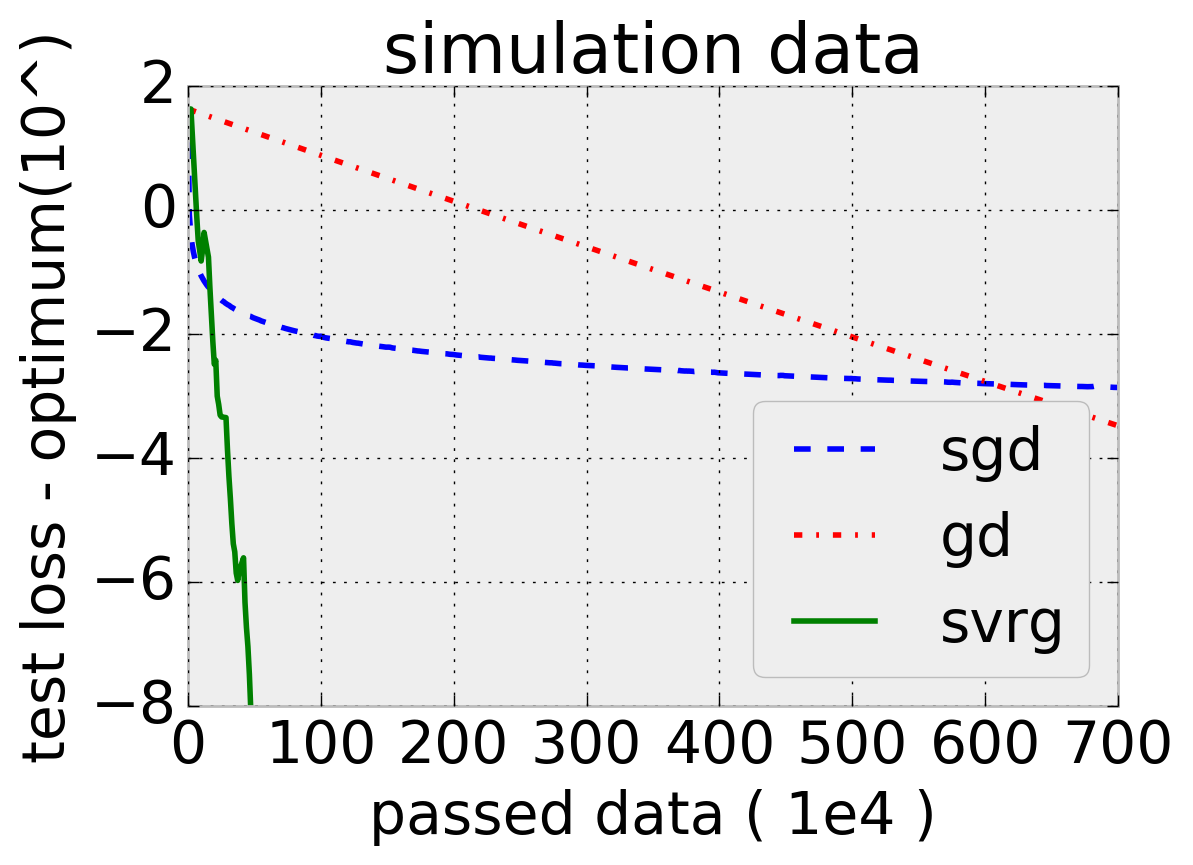}}
	\subfigure[]{
		\label{figf}
		\includegraphics[width=1.02in]{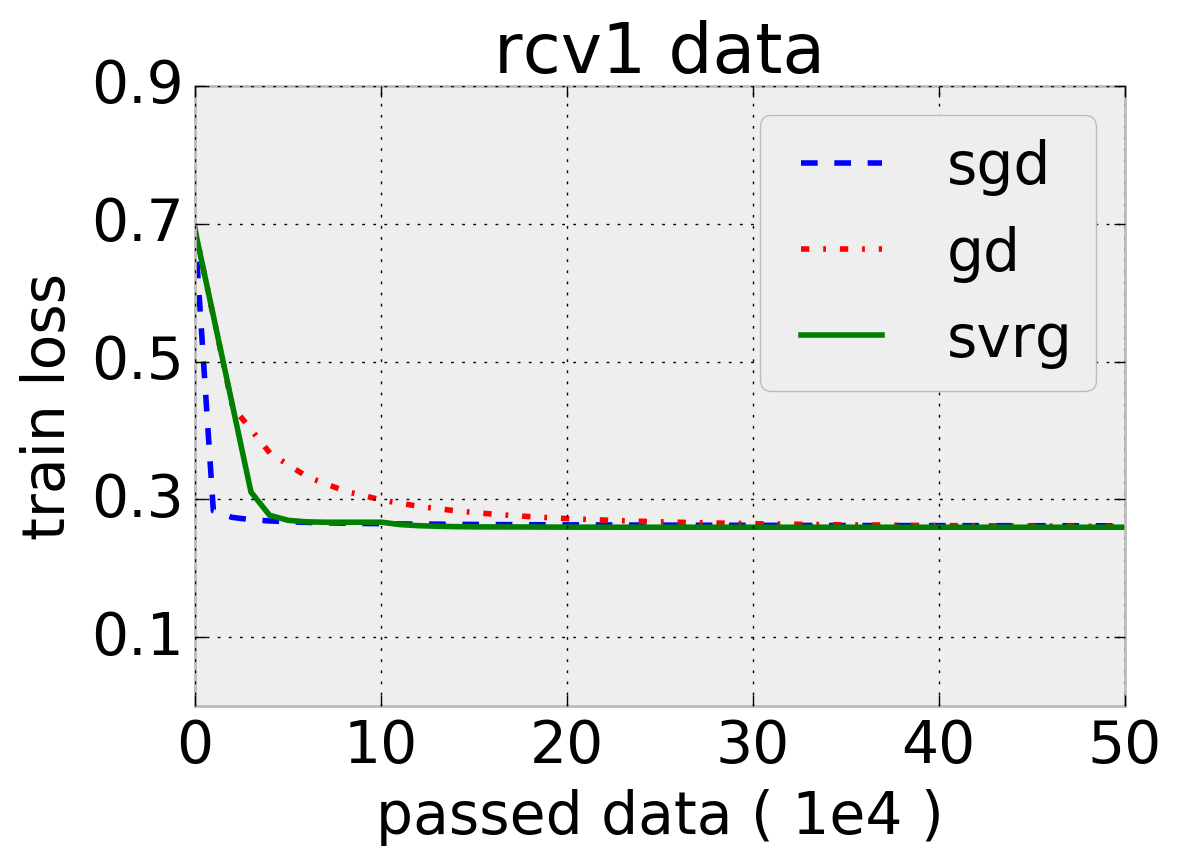}}
	\subfigure[]{
		\label{fige}
		\includegraphics[width=1.02in]{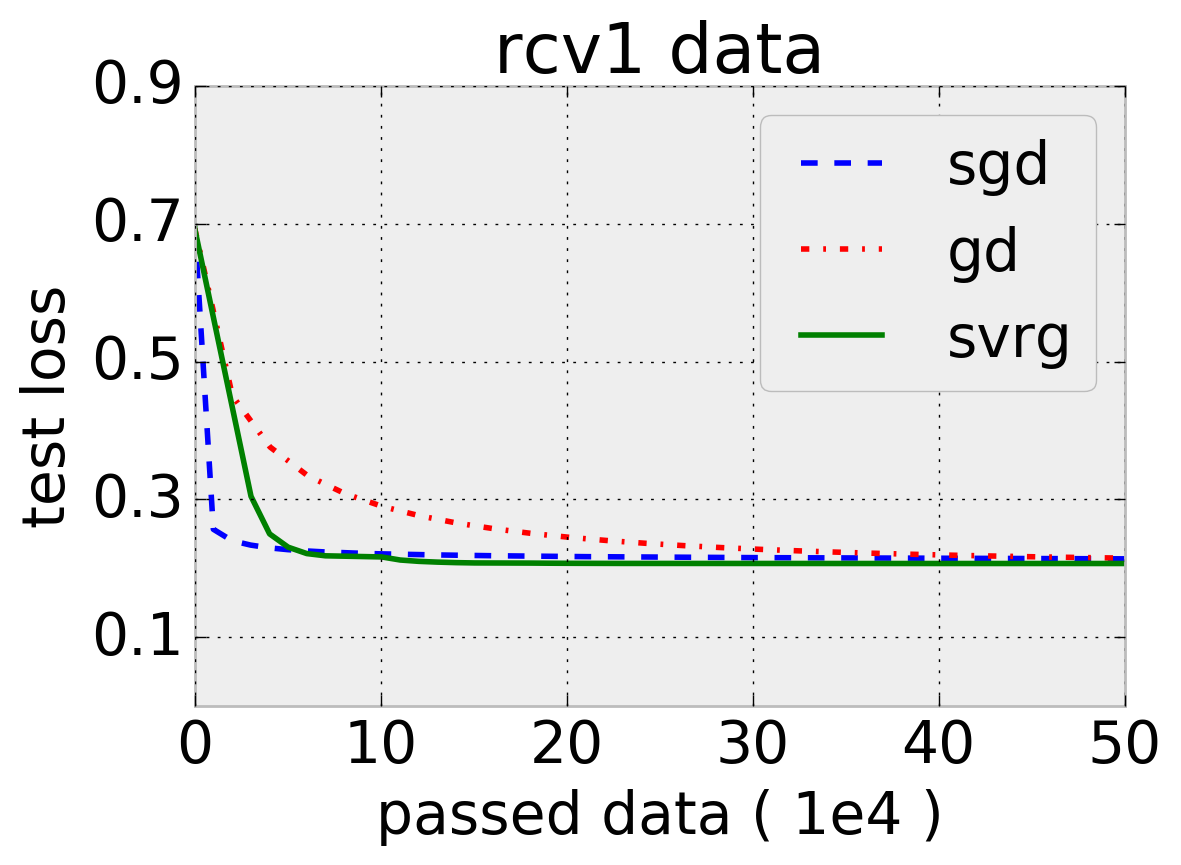}}
	\subfigure[]{
		\label{figd}
		\includegraphics[width=0.95in]{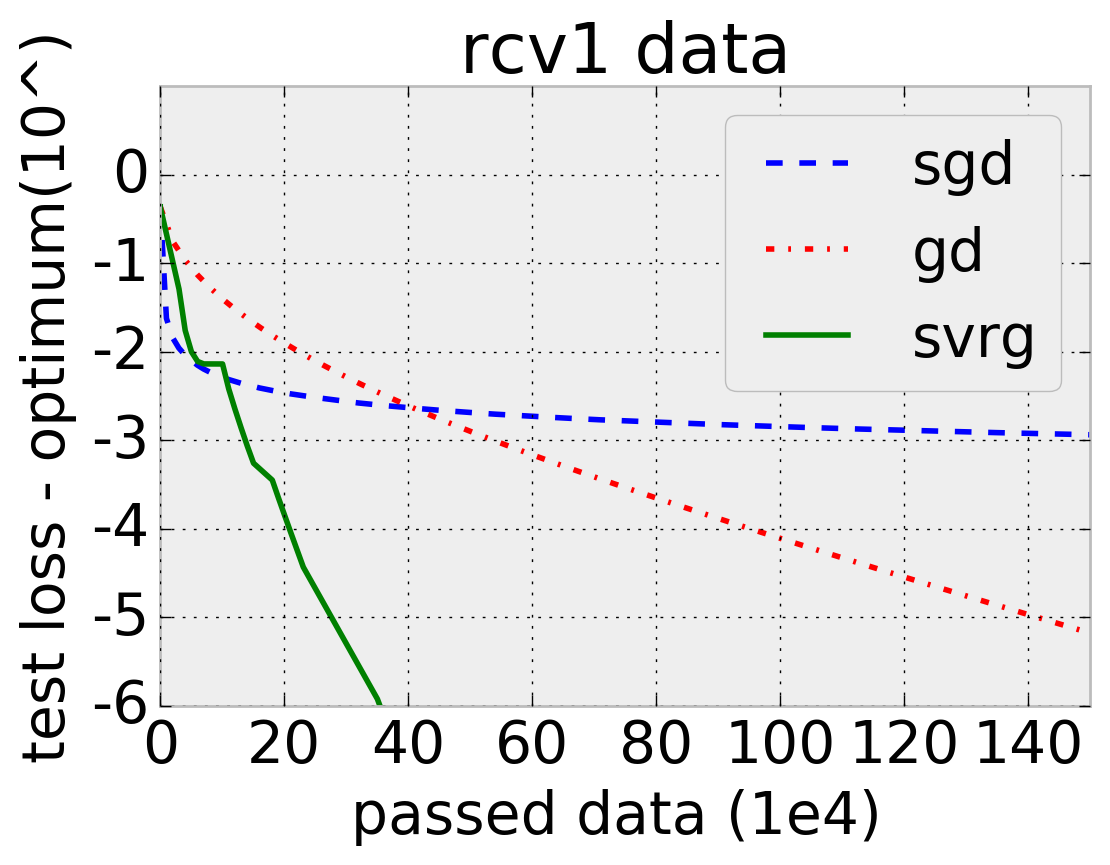}}
	\subfigure[]{
		\label{figh}
		\includegraphics[width=0.95in]{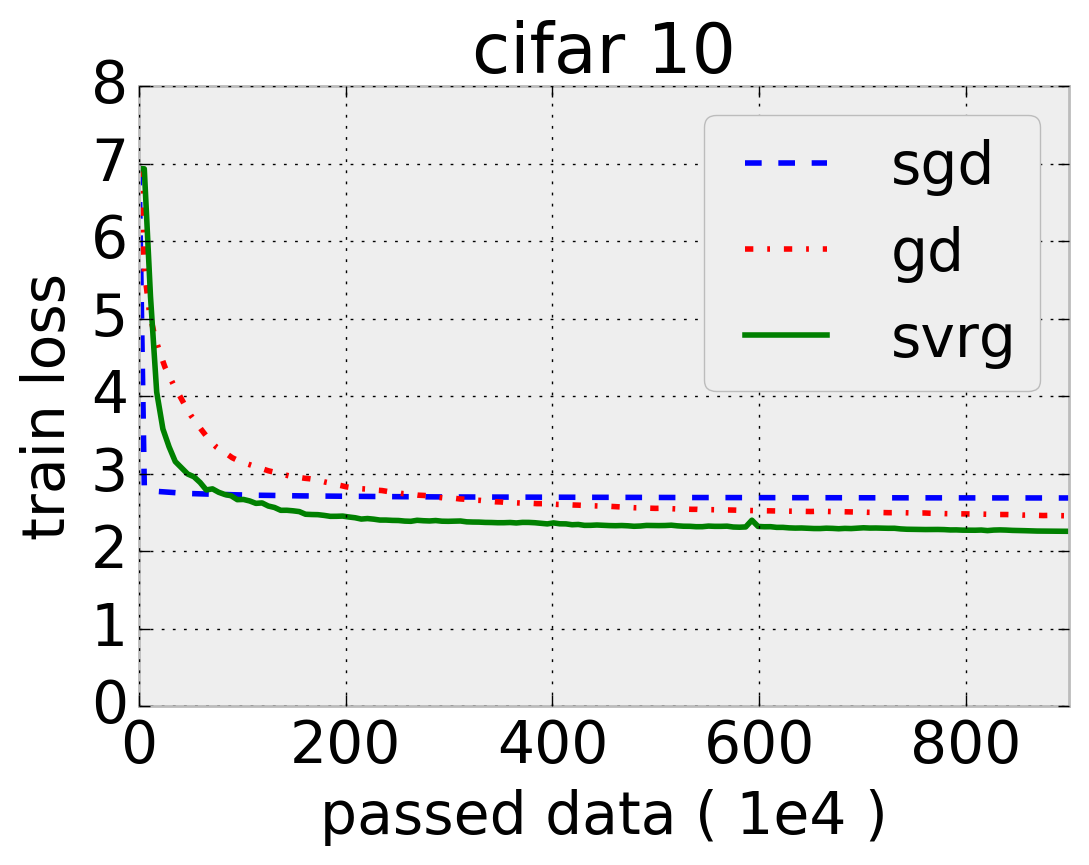}}
	\subfigure[]{
		\label{figi}
		\includegraphics[width=1.02in]{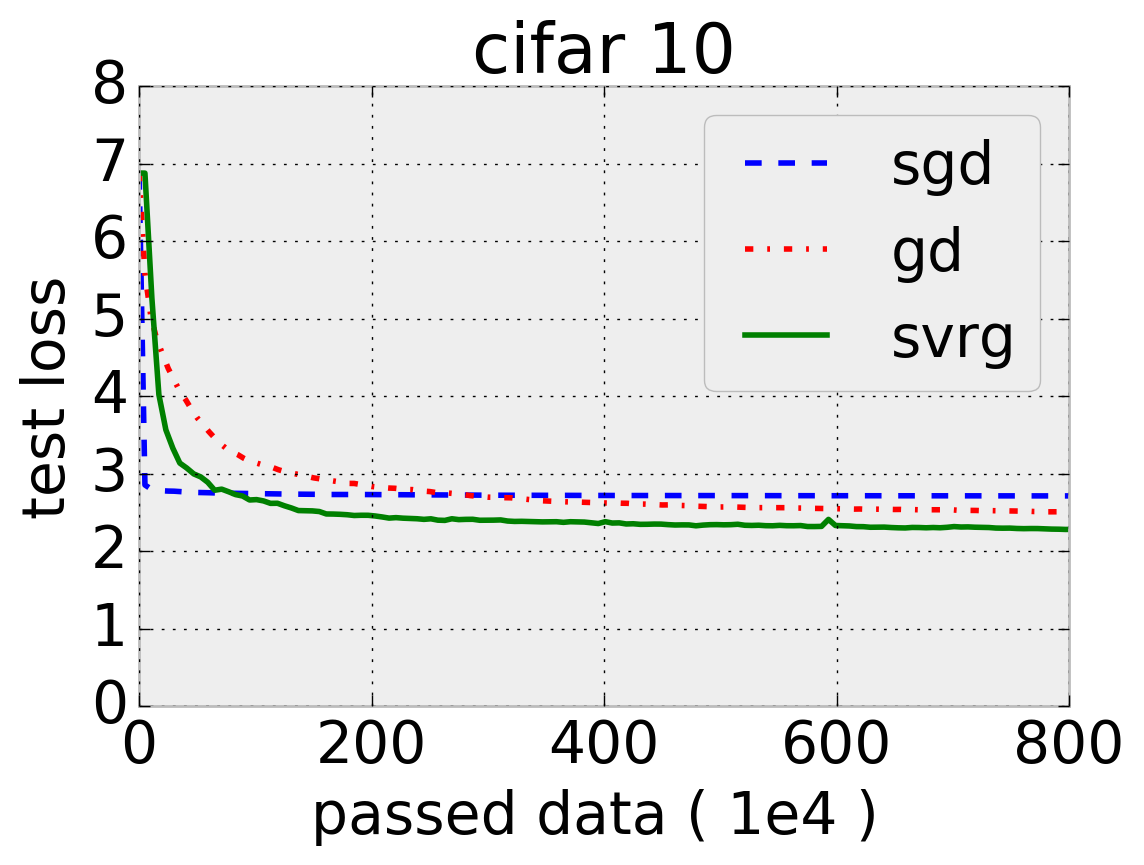}}
		\subfigure[]{
			\label{figg}
			\includegraphics[width=1.07in]{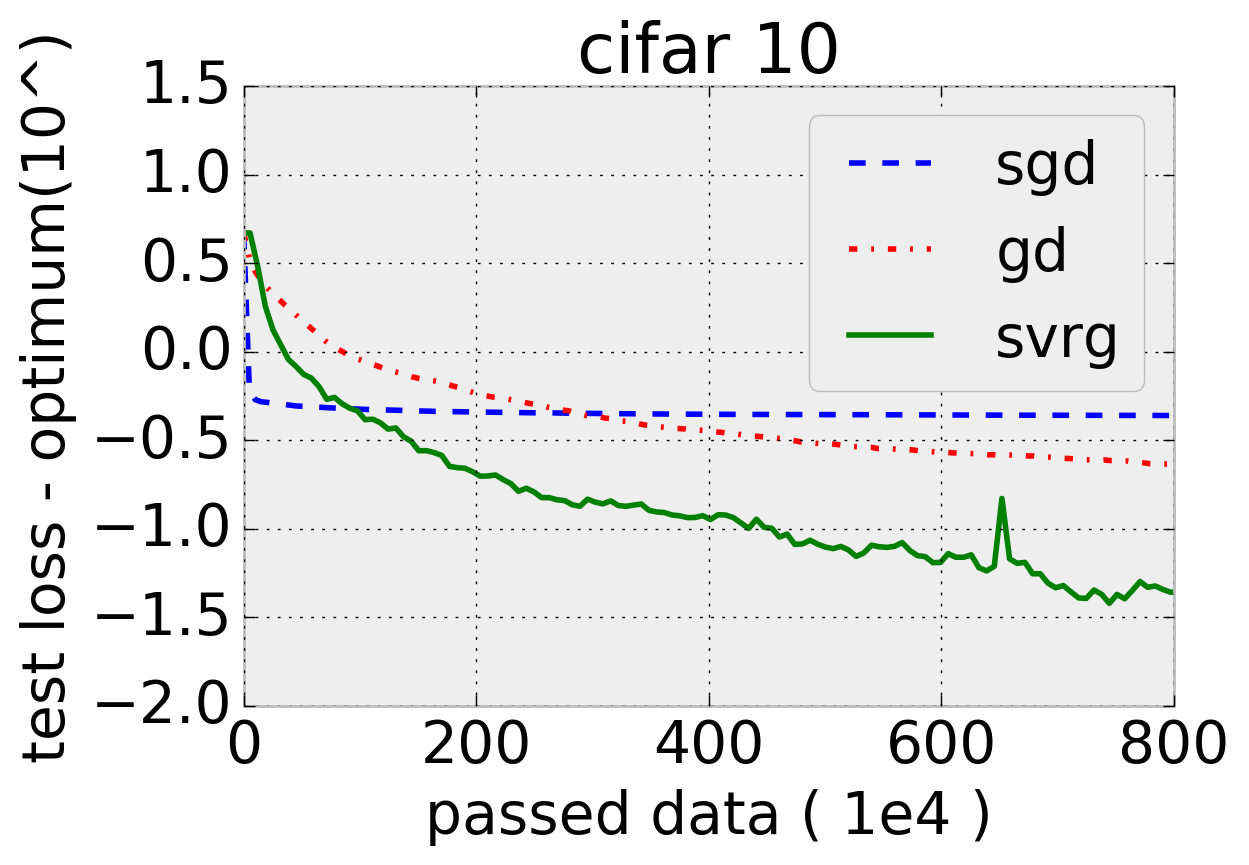}}
	\caption{Experimental Results}
	\label{Fig1}
\end{figure}

In this section, we report experimental results to validate our theoretical findings. We conducted experiments on three tasks: linear regression, logistic regression, and fully connected neural networks, whose objective functions are least square loss, logistic loss, and cross-entropy loss respectively, plus an $L_2$ regularization term with {\small$\lambda=1/\sqrt{n}$}. The first two tasks are used to verify our results for convex problems, and the third task is used to verify our theory on nonconvex problems. For each task, we report three figures. The horizontal axis of each figure corresponds to the number of data passes and the vertical axis corresponds to the training loss, test loss, and log-scaled test loss, respectively.
For linear regression, we independently sample data instances with size $n=40000$ from a $100-$dimension Gaussian distribution. We use half of them as the training data and the other as the test data.  We set the step size for GD, SGD, SVRG as $0.032$, $0.01/t$ and $0.005$, respectively, according to the smoothness and strong-convexity coefficients. For our simulated data, the condition number $\kappa\approx116$. The results are shown in Fig.\ref{figa}\ref{figb}\ref{figc}. For logistic regression, we conduct binary classification on benchmark dataset \textit{rcv1}. We set the step sizes for GD, SGD, SVRG as $400$, $200/t$ and $1$, respectively. The results are shown in Fig. \ref{figd}\ref{fige}\ref{figf}. For neural networks, we work on a model with one fully connected hidden layer of 100 nodes, ten softmax output nodes, and sigmoid activation \cite{johnson2013accelerating}. We tune the step size for GD, SGD, SVRG and eventually choose $0.03$, $0.25/\sqrt{t}$ and $0.001$, respectively, which correspond to the best performances in our experiments. The inner loop size for SVRG for convex problems is set as $2n$ and that for nonconvex problem is set as $5n$. The results are shown in Fig.\ref{figg}\ref{figh}\ref{figi}.

From the results for all the three tasks, we have the following observations. (1) As training error decreases, the test error also decreases. (2) According to Fig.\ref{figa}, SVRG is faster than GD by a factor of $\mathcal{O}(\kappa)$ and faster than SGD by a factor of more than $\mathcal{O}(\kappa)$. (3) According to Fig. \ref{figa}\ref{figd}\ref{figg}, SGD is the slowest although it is fast in the beginning, which is consistent with our discussions in Section \ref{sec4}.

By comparing the results of logistic regression and linear regression, we have the following observations. (1) The test error for logistic regression converges after fewer rounds of data passes than linear regression. This is because the condition number $\kappa$ for logistic regression is smaller than linear regression. (2) SVRG is faster than GD and SGD but the differences between them are less significant for logistic regression, due to a smaller $\kappa$.
As compared to the results for logistic regression and linear regression, we have the following observations on the results of neural networks. (1) The convergence rate is slower and the accuracy is lower. This is because of the nonconvexity and the gap between global optimum  and local optimum. (2) SVRG is faster than GD and SGD but the differences between them are not as significant as in the convex cases, which is consistent with our discussions in Section \ref{sec4} by considering the data size of CIFAR 10. 

\section{Conclusion}
In this paper, we have studied the generalization error bounds for optimization algorithms to solve R-ERM problems, by using stability as a tool. For convex problems, we have obtained both expected bounds and high-probability bounds. Some of our results can be extended to the nonconvex case. Roughly speaking, our theoretical analysis has shown: (1) Along with the training process, the generalization error will decrease; (2) SVRG outperforms GD and SGD in most cases. We have verified the theoretical findings by using experiments on linear regression, logistic regression and fully connected neural networks. In the future, we plan to study the stability of R-ERM with other regularization terms, e.g., the $L_1$ regularizer, which is usually associated with non-smooth optimization methods.
\bibliographystyle{aaai}
\bibliography{stability}
\newpage
\section{Appendices}
\subsection{Proofs of Lemma 3.2, Lemma 3.3 and Theorem 3.1}
\textbf{Lemma 3.2:}
\textit{For R-ERM problems, we have $\forall j\in\{1,\cdots,n\}$:
{\small\begin{equation}
	\mathbb{E}_S\left[R(f_{S,\mathcal{F}_c}^*)-R_{S}(f_{S,\mathcal{F}_c}^*)\right]=\mathbb{E}_S\left[l(f_{S,\mathcal{F}_c}^*,z'_j)-l(f_{S^{j},\mathcal{F}_c}^*,z'_j)\right]\label{eq7a}
	\end{equation}} and
{\small\begin{eqnarray}
	&&\mathbb{E}_S[\nabla R(f_{S,\mathcal{F}_c}^*)-\nabla R_{S}(f_{S,\mathcal{F}_c}^*)] \nonumber\\
	&=&\mathbb{E}_S[\nabla_fl(f_{S,\mathcal{F}_c}^*,z'_j)-\nabla_fl(f_{S^{j},\mathcal{F}_c}^*,z'_j)]\label{eq8a},
	\end{eqnarray}}
where $S^j=\{z_1,\cdots,z_{j-1},z'_j,z_{j+1},\cdots,z_n\}$, and $f_{S^j,r}^*$ is the minimizer of $R_{S^j}^r$.}
\vspace{3mm}

\textbf{Proof:}

The proofs of Eq.(\ref{eq7a}) and Eq.(\ref{eq8a}) are very similar, we only prove Eq.(\ref{eq8a}).
{\small\begin{eqnarray}
	&&\mathbb{E}_S[\nabla R_S(f_{S,\mathcal{F}_c}^*)]\\
	&=&\frac{1}{n}\sum_{j=1}^{n}\mathbb{E}_S[\nabla_fl(f_{S,\mathcal{F}_c}^*,z_j)] \\
	&=&\frac{1}{n}\sum_{j=1}^{n}\mathbb{E}_{S,z'_j}[\nabla_fl(f_{S,\mathcal{F}_c}^*,z_j)]\\
	&=&\frac{1}{n}\sum_{j=1}^{n}\mathbb{E}_{S,z'_j}[\nabla_fl(f_{S^{j},\mathcal{F}_c}^*,z'_j)] \label{eq1}
	\end{eqnarray}}
Using the definition of $R$, we can get
{\small\begin{equation}\label{eq2}
	\mathbb{E}_S\nabla R(f_{S,\mathcal{F}_c}^*)=\mathbb{E}_{S,z}\nabla_fl(f_{S,\mathcal{F}_c}^*,z)=\mathbb{E}_{S,z'_j}\nabla_fl(f_{S,\mathcal{F}_c}^*,z'_j).
	\end{equation}} By combining Eq.(\ref{eq1}) and (\ref{eq2}), we can get the results. $\quad\quad\Box$

\vspace{3mm}

\textbf{Lemma 3.3 :}\textit{
Assume that the loss function is $L$-Lipschitz and $\gamma$-smooth w.r.t. the prediction output vector, we have
{\small\begin{equation}
	\mathbb{E}_S[\nabla R(f_{S,\mathcal{F}_c}^*)-\nabla R_S(f_{S,\mathcal{F}_c}^*)]^2\leq\frac{L^2}{2n}+6L\gamma\beta_1.
	\end{equation}}}
\vspace{3mm}

\textbf{Proof:}

The proof is following Lemma 9 and Lemma 25 in \cite{bousquet2002stability}. We just need to relpalce $M$ which is the upper bound of the loss function by the upper bound of the derivative of loss function $\nabla_fl(f,z)$, and replace the lip By the assumption that the loss function is $L-$Lipschitz continous and $\gamma-$smooth w.r.t. the prediction output vector, we can get that $\nabla_fl(f,z)\leq L$ and $\nabla_fl(f,z)$ is $\gamma-$ Lipschitz continous.  Following Lemma 9 in \cite{bousquet2002stability}, we have
{\small\begin{eqnarray}
	&&\mathbb{E}_S[\nabla R(f_{S,\mathcal{F}_c}^*)-\nabla R_S(f_{S,\mathcal{F}_c}^*)]^2\\
	&\leq&\frac{L^2}{2n}+3L\mathbb{E}_{S,z'_j}[|\nabla l(f_{S,\mathcal{F}_c}^*,z_j)-\nabla l(f_{S^j,\mathcal{F}_c}^*,z_j)|]\\
	&\leq&\frac{L^2}{2n}+3L\gamma\mathbb{E}_{S,z'_j}[\|f_{S,\mathcal{F}_c}^*-f_{S^j,\mathcal{F}_c}^*\|_{\mathcal{F}_c}]\\
	&\leq&\frac{L^2}{2n}+3L\gamma\mathbb{E}_{S,z'_j}[\|f_{S,\mathcal{F}_c}^*-f_{S^{\setminus j},\mathcal{F}_c}^*\|_{\mathcal{F}_c}]\\
	&&+3L\gamma\mathbb{E}_{S^j,z'_j}[\|f_{S,\mathcal{F}_c}^*-f_{S^{\setminus j},\mathcal{F}_c}^*\|_{\mathcal{F}_c}]\\
	&\leq&\frac{L^2}{2n}+6L\gamma\beta_1.\quad\quad\Box
	\end{eqnarray}}

\textbf{Theorem 3.1 :}	\textit{	Consider an R-ERM problem, if the loss function is $L$-Lipschitz continuous, $\gamma$-smooth, and convex with respect to the prediction output vector, we have {\small\begin{eqnarray}
	\mathbb{E}_{S,A}\mathcal{E}&\leq&\mathcal{E}_{app}+2\beta_0+\mathbb{E}_{S,A}\rho_0(T)+\frac{\gamma\mathbb{E}_{S,A}\rho_1(T)}{2}\nonumber\\
	&&+\sqrt{\mathbb{E}_{S,A}\rho_1(T)\left(\frac{L^2}{2n}+6L\gamma\beta_1\right)},
	\end{eqnarray}}
where $\beta_0,\beta_1$ are the uniform stability and output stability of the R-ERM process as defined in Def.2.1 and Def.2.2, $\rho_0(T)$ and $\rho_1(T)$ are the convergence errors defined in Eqn.(16).}
\vspace{3mm}

\textbf{Proof:}\\
If $l(f,z)$ is convex and $\gamma$-smooth w.r.t. $f$,
we can get that $R(f)$ is $\gamma$-smooth and $R_S(f)$ is convex.
First, we decompose $\mathcal{E}_{opt}$ as follows:
{\small\begin{eqnarray}
	&&\mathcal{E}_{opt}\\
	&=& R(f_T)-R(f_{S,\mathcal{F}_c}^*)\\
	&\leq&\nabla R(f_{S,\mathcal{F}_c}^*)^T(f_T-f_{S,\mathcal{F}_c}^*)+\frac{\gamma}{2}\|f_T-f_{S,\mathcal{F}_c}^*\|_{\mathcal{F}_c}^2\\
	&=&\left(\nabla R(f_{S,\mathcal{F}_c}^*)-\nabla R_{S}(f_{S,\mathcal{F}_c}^*)\right)^T(f_T-f_{S,\mathcal{F}_c}^*)\\
	&\quad&+\nabla R_{S}(f_{S,\mathcal{F}_c}^*)^T(f_T-f_{S,\mathcal{F}_c}^*)+\frac{\gamma}{2}\|f_T-f_{S,\mathcal{F}_c}^*\|_{\mathcal{F}_c}^2 \\
	&\leq&\left(\nabla R(f_{S,\mathcal{F}_c}^*)-\nabla R_{S}(f_{S,\mathcal{F}_c}^*)\right)^T(f_T-f_{S,\mathcal{F}_c}^*) \label{eq13}\\
	&\quad&+R_{S}(f_T)-R_{S}(f_{S,\mathcal{F}_c}^*)+\frac{\gamma}{2}\|f_T-f_{S,\mathcal{F}_c}^*\|_{\mathcal{F}_c}^2, \nonumber
	\end{eqnarray}}
where the first inequality is established by using the $\gamma$-smoothness condition and the third inequality is established by using the convexity condition.
Taking expectation w.r.t. $S$ and the optimization algorithm $A$, we can get
{\small\begin{eqnarray}
	\mathbb{E}_{S,A}\frac{\gamma}{2}\|f_T-f_{S,\mathcal{F}_c}^*\|^2&=&\frac{\gamma}{2}\mathbb{E}_{S,A}\rho_1(T)\\
	\mathbb{E}_{S,A}[R_{S}(f_T)-R_{S}(f_{S,\mathcal{F}_c}^*)]&=&\mathbb{E}_{S,A}\rho_0(T)
	\end{eqnarray}}
For the term {\small$\left(\nabla R(f_{S,\mathcal{F}_c}^*)-\nabla R_{S}(f_{S,\mathcal{F}_c}^*)\right)^T(f_T-f_{S,\mathcal{F}_c}^*)$}, by using Cauthy-Schwarz inequality, we can get:
{\small\begin{eqnarray}
	&&\quad\mathbb{E}_{S,A}\left(\nabla R(f_{S,\mathcal{F}_c}^*)-\nabla R_S(f_{S,\mathcal{F}_c}^*)\right)^T(f_T-f_{S,\mathcal{F}_c}^*)\nonumber\\
	&&\leq\sqrt{\mathbb{E}_{S,A}\|f_T-f_{S,\mathcal{F}_c}^*\|^2\mathbb{E}_S\left(\nabla R(f_{S,\mathcal{F}_c}^*)-\nabla R_S(f_{S,\mathcal{F}_c}^*)\right)^2}\nonumber\\
	&&\leq\sqrt{\mathbb{E}_{S,A}\rho_1(T)\left(\frac{L^2}{2n}+6L\gamma\beta_1\right)}\label{ineq1}
	\end{eqnarray}}
where the second inequality holds according to Lemma 3.3.

Next we decompose $\mathcal{E}_{est}$ as follows:
{\small\begin{eqnarray}
	&&\mathcal{E}_{est}\\
	&=&\left[R(f_{S,\mathcal{F}_c}^*)-R_S(f_{S,\mathcal{F}_c}^*)\right]+\left[R_S(f_{S,\mathcal{F}_c}^*)-R_S(f_{\mathcal{F}_c}^*)\right]\nonumber\\
	&\quad&+\left[R_S(f_{\mathcal{F}_c}^*)-R(f_{\mathcal{F}_c}^*)\right]\\
	&\leq&\left[R(f_{S,\mathcal{F}_c}^*)-R_S(f_{S,\mathcal{F}_c}^*)\right]+\left[R_S(f_{\mathcal{F}_c}^*)-R(f_{\mathcal{F}_c}^*)\right] \label{eq17},
	\end{eqnarray}}
where the second inequality is established because $f_{S,\mathcal{F}_c}^*$ is the minimizer of $R_S$ restricted to the hypothesis space $\mathcal{F}_c$.

Since {\small$f_{\mathcal{F}_c}^{*}$} is independent of $S$, we have {\small$\mathbb{E}_S\left[R_S(f_{\mathcal{F}_c}^*)-R(f_{\mathcal{F}_c}^*)\right]=0$}. Then by using Eq.(\ref{eq7}) and the definition of uniform stability, we can get
{\small\begin{eqnarray}
	\mathbb{E}_S\mathcal{E}_{est}&\leq&\mathbb{E}_S\left[R(f_{S,\mathcal{F}_c}^*)-R_S(f_{S,\mathcal{F}_c}^*)\right] \nonumber\\
	&\leq&\mathbb{E}_{S,z'_j}[|l(f_{S,\mathcal{F}_c}^*,z'_j)-l(f_{S^j,\mathcal{F}_c}^*,z'_j)|] \nonumber\\
	&\leq&\mathbb{E}_{S,z'_j}[|l(f_{S,\mathcal{F}_c}^*,z'_j)-l(f_{S^{\setminus j},\mathcal{F}_c}^*,z'_j)|] \nonumber\\
	&&+\mathbb{E}_{S,z'_j}[|l(f_{S^{\setminus j},\mathcal{F}_c}^*,z'_j)-l(f_{S^j,\mathcal{F}_c}^*,z'_j)|] \nonumber\\
	&\leq& 2\beta_0. \label{ineq2}
	\end{eqnarray}}
By combining Ineq.(\ref{ineq1}) and Ineq.(\ref{ineq2}), we can get the result in the theorem.
$\quad\quad\Box$
\vspace{3mm}

\subsection{Proof of Theorem 3.5}	
\vspace{3mm}
\textbf{Theorem 3.5:}\textit{
For an R-ERM problem, if the loss function is L-Lipschitz continuous, $\gamma$-smooth and convex with respect to the prediction output vector, and $0\leq l(f_{S,\mathcal{F}_c}^*,z)\leq M$ for arbitrary $z\in\mathcal{Z}$ and $S\in\mathcal{Z}^n$, then with probability at least $1-\delta$, we have
{\small\begin{eqnarray*}
		\mathcal{E}&\leq&\mathcal{E}_{app}+2\beta_0+\rho_0(T)+\frac{\gamma}{2}\rho_1(T)+2\gamma\beta_1\sqrt{\rho_1(T)}\\	
		&&+\left(4n\beta_0+2M+(4n\gamma\beta_1+L)\sqrt{\rho_1(T)}\right)\sqrt{\frac{\ln{4/\delta}}{2n}}.
	\end{eqnarray*}}	}
	In order to prove Theorem 3.5, we need to use the following theorem  which is proposed by McDiarmid.
	\vspace{3mm}
	
	\textbf{Theorem} \textit{(McDiarmid,1989):
	Let $S$ and $S^i$ are two data sets which are different at only one point $j$. Let $F: \mathcal{Z}^n\rightarrow\mathcal{R}$ be any measurable function which there exists constants $c_j(j=1,\cdots,n)$ such that {\small$sup_{S\in\mathcal{Z}^n,z'_j\in\mathcal{Z}}|F(S)-F(S^j)|\leq c_j$}, then {\small$P_S(F(S)-\mathbb{E}_SF(S)\geq\epsilon)\leq e^{-2\epsilon^2/\sum_{j=1}^nc_j^2}$}.}
	\vspace{3mm}
	
	\textbf{Proof:}
	
	Firstly, we give the high probability bound for estimation error $\mathcal{E}_{est}$. As we have the decomposition Ineq. (\ref{eq17}),  we need to analyze {\small$R(f_{S,\mathcal{F}_c}^*)-R_S(f_{S,\mathcal{F}_c}^*)$} and {\small$R_S(f_{\mathcal{F}_c}^*)-R(f_{\mathcal{F}_c}^*)$}.
	By using Theorem 12 in \cite{bousquet2002stability}, we have with probability at least $1-\delta$,
	{\small\begin{equation}\label{ineq3}
		R(f_{S,\mathcal{F}_c}^*)-R_S(f_{S,\mathcal{F}_c}^*)\leq2\beta_0+(4m\beta_0+M)\sqrt{\frac{\ln{1/\delta}}{2n}}.
		\end{equation}}
	For {\small$R_S(f_{\mathcal{F}_c}^*)-R(f_{\mathcal{F}_c}^*)$}, by using Hoeffding's inequality, we can get with probability at least $1-\delta$,
	{\small\begin{equation}\label{ineq4}
		R_S(f_{\mathcal{F}_c}^*)-R(f_{\mathcal{F}_c}^*)\leq M\sqrt{\frac{\ln{1/\delta}}{2n}}
		\end{equation}}	
	We can use Hoeffding's bound since $f_{\mathcal{F}_c}^*$ is independent with the training set $S$.
	By combining Ineq.(\ref{ineq3}) and Ineq.(\ref{ineq4}), we have with probability at least $1-2\delta$,
	{\small\begin{equation}\label{eq6}
		\mathcal{E}_{est}\leq 2\beta_0+(4m\beta_0+2M)\sqrt{\frac{\ln{1/\delta}}{2n}}.
		\end{equation}}
	Secondly, we give high probability bound for the term $\mathcal{E}_{opt}$ by using Theorem (McDiarmid, 1989). According to Theorem (McDiarmid, 1989), we need to calculate {\small$\mathbb{E}_S[\nabla R(f_{S,\mathcal{F}_c}^*)-\nabla R_S(f_{S,\mathcal{F}_c}^*)]$}, and {\small$c_j=|\nabla R(f_{S,\mathcal{F}_c}^*)-\nabla R_S(f_{S,\mathcal{F}_c}^*)-\left(\nabla R(f_{S^{j},\mathcal{F}_c}^*)-\nabla R_S(f_{S^{j},\mathcal{F}_c}^*)\right)|$}.
	
	For {\small$\nabla R(f_{S,\mathcal{F}_c}^*)-\nabla R_S(f_{S,\mathcal{F}_c}^*)$}, we have
	{\small\begin{eqnarray}
		&&\mathbb{E}_S[\nabla R(f_{S,\mathcal{F}_c}^*)-\nabla R_S(f_{S,\mathcal{F}_c}^*)]\\
		&=&\mathbb{E}_S\left[\nabla_fl(f_{S,\mathcal{F}_c}^*,z'_j)-\nabla_fl(f_{S^{j},\mathcal{F}_c}^*,z'_j)\right]\leq2\gamma\beta_1.
		\end{eqnarray}}
	We also have
	{\small\begin{eqnarray}
		&&|\nabla R(f_{S,\mathcal{F}_c}^*)-\nabla R(f_{S^{\setminus j,r}}^*)|\\
		&\leq& \mathbb{E}_S\left[|\nabla_fl(f_{S,\mathcal{F}_c}^*,z'_j)-\nabla_fl(f_{S^{j},\mathcal{F}_c}^*,z'_j)|\right]\nonumber\\
		&\leq&\gamma|f_{S,\mathcal{F}_c}^*(x)-f_{S^{\setminus j},\mathcal{F}_c}^*(x)|\leq\gamma\beta_1
		\end{eqnarray}}
	and
	{\small
		$|\nabla R_S(f_{S,\mathcal{F}_c}^*)-\nabla R_S(f_{S^{\setminus j},\mathcal{F}_c}^*)|\leq \gamma\beta_1+\frac{L}{n},$
	}
	which yields
	{\small\begin{eqnarray}
		|\nabla R(f_{S,\mathcal{F}_c}^*)-\nabla R(f_{S^{j},\mathcal{F}_c}^*)|&\leq& 2\gamma\beta_1\\
		|\nabla R_S(f_{S,\mathcal{F}_c}^*)-\nabla R_S(f_{S^{j},\mathcal{F}_c}^*)|&\leq&2\gamma\beta_1+\frac{L}{n}.
		\end{eqnarray}}
	Thus we can get $c_j=4\gamma\beta_1+\frac{L}{n}$.
	By using Theorem (McDiarmid, 1989), we can get that with probability at least $1-\delta$
	{\small\begin{equation}\label{eq5}
		\nabla R(f_{S,\mathcal{F}_c}^*)-\nabla R_S(f_{S,\mathcal{F}_c}^*)\leq 2\gamma\beta_1+(4n\gamma\beta_1+L)\sqrt{\frac{\ln 1/\delta}{2n}}.
		\end{equation}}
	By putting Ineq.(\ref{eq5}), $\rho_0(T)$ and $\rho_1(T)$ in Ineq.(\ref{eq13}), we have with probability at least $1-2\delta$,
	{\small\begin{eqnarray}
		\mathcal{E}_{opt}&\leq&\left(2\gamma\beta_1+(4n\gamma\beta_1+L)\sqrt{\frac{\ln 1/\delta}{2n}}\right)\sqrt{\rho_1(T)}\nonumber\\
		&&+\rho_0(T)+\left(\frac{\gamma}{2}\right)\rho_1(T)\label{eq9}
		\end{eqnarray}}
	By combining Ineq.(\ref{eq6}) and Ineq.(\ref{eq9}), we have with probability $1-4\delta$,
	{\small\begin{eqnarray*}
			&&\mathcal{E}\leq\mathcal{E}_{app}+2\beta_0+(4n\beta_0+2M)\sqrt{\frac{\ln{1/\delta}}{2n}}+\rho_0(T)\\
			&&+\left(\frac{\gamma}{2}\right)\rho_1(T)+\left(2\gamma\beta_1+(4n\gamma\beta_1+L)\sqrt{\frac{\ln 1/\delta}{2n}}\right)\sqrt{\rho_1(T)}\\
			&&\leq\mathcal{E}_{app}+2\beta_0+\rho_0(T)+\left(\frac{\gamma}{2}\right)\rho_1(T)+2\gamma\beta_1\sqrt{\rho_1(T)}\\
			&&+\left(4n\beta_0+2M+(4n\gamma\beta_1+L)\sqrt{\rho_1(T)}\right)\sqrt{\frac{\ln{1/\delta}}{2n}}
		\end{eqnarray*}}
		By using $\delta/4$ to replace $\delta$, we can get the result.
		$\Box$
		\subsection{Proof of Theorem 3.9}
		\vspace{3mm}
		
		\textbf{Theorem 3.9:}\textit{
		If $R_{S}^r$ is $\mu$-strongly convex in the $\epsilon_0$- neighborhood of arbitrary local minimum $w_{loc}$, satisfies strict saddle point property, $L$- Lipschitz continuous, $\gamma$-smooth and continuously twice differential w.r.t the model parameter $w$, and the loss function is convex w.r.t $f$, then we have
		{\small\begin{equation}
			\mathbb{E}_{S,A}\mathcal{E}\leq\mathcal{E}_{app}+2\beta_0+R(w_{loc})-R(w_{S,\mathcal{F}_c}^*)+\frac{L}{\mu}\sqrt{\min_{t=1,\cdots,T}\mathbb{E}_{S,A}\rho_2(t)},
			\end{equation}}
		where $T\geq T_1$ and $T_1$ is the number of iterations to achieve  {\small$\min_{t=1,\cdots,T_1}\mathbb{E}_{S,A}\left[\rho_2(t)\right]\leq\gamma^2\epsilon_0^2$}.}
		\vspace{3mm}
		
		\textbf{Proof:}
		
		The upper bound for expected estimation error is the same as convex cases since the loss function is convex w.r.t $f$, i.e., $\mathbb{E}_{S}\mathcal{E}_{est}\leq2\beta_0$.
		
		Referring to a recent work of Lee \textit{et.al} \cite{lee2016gradient}, GD with a random initialization and sufficiently small constant step size converges to a local minimizer almost surely under the assumptions in Theorem 3.9. Thus, the assumption that $R_{S}^r$ is $\mu$-strongly convex in the $\epsilon_0$-neighborhood of arbitrary local minimum $w_{loc}$ is easily to be satisfied in sense of "almost surely".  We decompose $\min_{t=1,\cdots,T}\mathbb{E}\mathcal{E}_{opt}$ as {\small\begin{equation}\min_{t=1,\cdots,T}\mathbb{E}[R(w_t)-R(w_{loc})]+\mathbb{E}[R(w_{loc})-R(w_{S,\mathcal{F}_c}^*)].\end{equation}} By the L-Lipschitz condition, we have {\small$R(w_t)-R(w_{loc})\leq L\|w_t-w_{loc}\|$}. Firstly,
		We need to calculate how many iterations are needed to guarantee that
		{\small\begin{equation}\label{eq10}
			\min_{t=1,\cdots,T_1}\mathbb{E}\|w_t-w_{loc}\|\leq\epsilon_0.
			\end{equation}}
		By the $\gamma$-smooth assumption, we have {\small$\gamma\|w_t-w_{loc}\|^2\geq\langle\nabla R_{S}^r(w_t),w_t-w_{loc}\rangle$}. Thus for $w_t\in \mathcal{B}(w_{loc},\epsilon_0)$, we have {\small$\|\nabla R_{S}^r(w_t)\|\leq \gamma\|w_t-w_{loc}\|\leq \gamma\epsilon_0.$} By the continuously twice differential assumption, we can assume that {\small$\|\nabla R_{S}^r(w_t)\|\leq \gamma\epsilon_0$} for $w_t\in \mathcal{B}(w_{loc},\epsilon_0)$ and {\small$\|\nabla R_{S}^r(w_t)\|\geq \gamma\epsilon_0$} for $w_t\notin \mathcal{B}(w_{loc},\epsilon_0)$ without loss of generality. \footnote{Otherwise, we can choose $\epsilon_0$ small enough to make it satisfied.} Therefore {\small$	\min_{t=1,\cdots,T_1}\mathbb{E}\|\nabla R_{S}^r(w_t)\|^2\leq \gamma^2\epsilon_0^2$} is a sufficient condition for {\small$\min_{t=1,\cdots,T_1}\mathbb{E}\|w_t-w_{loc}\|\leq\epsilon_0$}.
		
		If $T\geq T_1$, by the $\mu$-strongly convex assumption, we have {\small$\|w_t-w_{loc}\|^2\leq\frac{1}{\mu}\langle\nabla R_{S}^r(w_t),w_T-w_{loc}\rangle\leq\frac{1}{\mu}\|\nabla R_{S}^r(w_t)\|\|w_t-w_{loc}\|$ for $w_t\in \mathcal{B}(w_{loc},\epsilon_0)$}, which yields {\small$\|w_t-w_{loc}\|\leq\frac{1}{\mu}\|\nabla R_{S}^r(w_t)\|.$}
		
		Based on the above discussions, we can get
		{\small\begin{eqnarray*}
				&&\min_{t=1,\cdots,T}\mathbb{E}\mathcal{E}_{opt}\\
				&=&\min_{t=1,\cdots,T}\mathbb{E}[R(w_t)-R(w_{loc})]+\mathbb{E}[R(w_{loc})-R(w_{S,\mathcal{F}_c^*})]\\
				&\leq& L\min_{t=1,\cdots,T}\mathbb{E}\|w_t-w_{loc}\|+\mathbb{E}[R(w_{loc})-R(w_{S,\mathcal{F}_c^*})]\\
				&\leq& \frac{L}{\mu}\min_{t=1,\cdots,T}\mathbb{E}\|\nabla R_S^r(w_t)\|+\mathbb{E}[R(w_{loc})-R(w_{S,\mathcal{F}_c^*})]\\
				&\leq& \frac{L}{\mu}\sqrt{\min_{t=1,\cdots,T}\mathbb{E}\|\nabla R_S^r(w_t)\|^2}+\mathbb{E}[R(w_{loc})-R(w_{S,\mathcal{F}_c^*})]\\
				&=& \frac{L}{\mu}\sqrt{\min_{t=1,\cdots,T}\mathbb{E}\rho_2(t)}+\mathbb{E}[R(w_{loc})-R(w_{S,\mathcal{F}_c^*})],
			\end{eqnarray*}}
			where $T\geq T_1$.
			$\Box$	
\end{document}